\pdfoutput=1

\documentclass[11pt]{article}

\usepackage[]{acl}

\usepackage{times}
\usepackage{latexsym}

\usepackage[T1]{fontenc}

\usepackage[utf8]{inputenc}

\usepackage{microtype}

\makeatletter
\newcommand\smaller{\@setfontsize\smaller{10}{5}}
\makeatother

\usepackage{multirow}
\usepackage{todonotes}

\usepackage{tablefootnote}
\usepackage{color, colortbl}
\definecolor{Gray}{gray}{0.93}

\usepackage{soul}

\usepackage{tabularx,booktabs}
\newcolumntype{Y}{>{\centering\arraybackslash}X}

\title{Efficient Classification of Long Documents Using {T}ransformers}

\author{Hyunji Hayley Park\\
  University of Illinois\thanks{\textsuperscript{ } Work done while at Amazon} \\
  {\small\texttt{hpark129@illinois.edu}} \\\And
  Yogarshi Vyas\\
  AWS AI Labs\\
  {\small\texttt{yogarshi@amazon.com}} \\\And
  Kashif Shah\\
  Microsoft$^{*}$\\
  {\small\texttt{kashifshah@microsoft.com}}}

\begin{document}
\maketitle
\begin{abstract}
Several methods have been proposed for classifying long textual documents using Transformers. However, there is a lack of consensus on a benchmark to enable a fair comparison among different approaches. 
In this paper, we provide a comprehensive evaluation of the relative efficacy measured against various baselines and diverse datasets --- both in terms of accuracy as well as time and space overheads. Our datasets cover binary, multi-class, and multi-label classification tasks and represent various ways information is organized in a long text (e.g. information that is critical to making the classification decision is at the beginning or toward the end of the document). Our results show that more complex models often fail to outperform simple baselines and yield inconsistent performance across datasets.
These findings emphasize the need for future studies to consider comprehensive baselines and datasets that better represent the task of long document classification to develop robust models.\footnote{Our code is available at\\\url{https://github.com/amazon-research/efficient-longdoc-classification}.}
\end{abstract}

\section{Introduction}
\label{intro}

Transformer-based models \citep{vaswani2017attention} have achieved much progress across many areas of NLP including text classification \cite{Minaee_2021}.
However, such progress is often limited to short sequences because self-attention requires quadratic computational time and space with respect to the input sequence length.
Widely-used models like BERT \citep{devlin-etal-2019-bert} or RoBERTa \citep{liu2019roberta} are typically pretrained to process up to 512 tokens.
This is problematic because real-world data can be arbitrarily long. As such, different models and strategies have been proposed to process longer sequences.

In particular, we can identify a few standard approaches for the task of long document classification.
The simplest approach is to truncate long documents --- using BERT or RoBERTa on the first 512 tokens is often used as a baseline.
More efficient Transformer models like Longformer \cite{beltagy2020longformer} and Big Bird \cite{zaheer2020big} use sparse self-attention instead of full self-attention to process longer  documents (e.g. up to 4,096 tokens). 
Other approaches process long documents in their entirety by dividing them into smaller chunks \cite[e.g.][]{pappagari2019hierarchical}. An alternative idea proposed by recent work is to select sentences from the document that are salient to making the classification decision \cite{ding_cogltx_2020}. 

However, the relative efficacy of these models is not very clear due to a lack of consensus on benchmark datasets and baselines.
\citet{tay2020long} propose a benchmark for comparing Transformers that can operate over long sequences, but this only includes a single, simulated\footnote{The benchmark considers the task of classifying IMDB reviews \cite{maas-etal-2011-learning} using byte-level information to simulate longer documents.} long document classification task.
Novel variants of efficient Transformers are often compared to a BERT/RoBERTa baseline only, without much comparison to other Transformer models designed for the task \cite[e.g.][]{beltagy2020longformer, zaheer2020big}.
Conversely, models designed for long document classification often focus exclusively on state-of-the-art models for particular datasets, and do not consider a BERT/RoBERTa baseline or any other Transformer models \cite[e.g.][]{ding_cogltx_2020,pappagari2019hierarchical}.

This paper provides a much-needed comprehensive comparison among existing models for long document classification by evaluating them against unified datasets and baselines. 
We compare models that represent different approaches on various datasets and against Transformer baselines. 
Our datasets cover binary, multi-class, and multi-label classification. We also consider different ways information that is relevant to the classification is organized in texts (e.g. in the beginning or toward the end) and how this affects model performance.
We also compare the models in terms of their training time, inference time, and GPU memory requirements to account for additional complexity that some of the models have relative to a BERT baseline. This allows us to compare the practical efficacy of the models for real-world usage. 

Our results show that more sophisticated models are often outperformed by simpler models (often including a BERT baseline) and yield inconsistent performance across datasets. 
Based on these findings, we  highlight the importance of considering diverse datasets while developing models, especially those that represent different ways key information is presented in long texts.
Additionally, we recommend that future research should also always include simpler baseline models. To summarize, our contributions are:
\begin{itemize}
    \item We provide insights into the practical efficacy of existing models for long document classification by evaluating them across different datasets, and against several baselines. We compare the accuracy of these models as well as their runtime and memory requirements.
    \item We present a comprehensive suite of evaluation datasets for long document classification with various data settings for future studies.
    \item We propose simple models that often outperform complex models and can be challenging baselines for future models for this task.
\end{itemize}

\section{Methods}

In this paper, we compare models representing different approaches to long document classification \cite{beltagy2020longformer, pappagari2019hierarchical, ding_cogltx_2020} on unified datasets and baselines. 

\subsection{Existing Models} 

As described in \S\ref{intro}, four distinct approaches have been proposed for long document classification: 1) document truncation, 2) efficient self-attention, 3) chunk representations, 4) key sentence selection.
We evaluate a representative model from each category in this work.

\paragraph{BERT (document truncation)} The simplest approach consists of finetuning BERT after truncating long documents to the first 512 tokens.\footnote{In practice, the first 510 tokens are used along with the [CLS] and [SEP] tokens. We use the token count including the two special tokens throughout the paper for simplicity.} 
As in \citet{devlin-etal-2019-bert}, we use a fully-connected layer on the [CLS] token for classification.
This is an essential baseline  as it establishes the limitations of a vanilla BERT model in classifying long documents yet is still competitive \cite[e.g.][]{beltagy2020longformer,chalkidis-etal-2019-large}. However, some prior work fails to consider this baseline \cite[e.g.][]{ding_cogltx_2020, pappagari2019hierarchical}.

\paragraph{Longformer (efficient self-attention)} We select Longformer \cite{beltagy2020longformer} as a model designed to process longer input sequences based on efficient self-attention that scales linearly with the length of the input sequence \cite[see][for a detailed survey]{tay2020efficient}. 
Longformer also truncates the input, but it has the capacity to process up to 4,096 tokens rather than 512 tokens as in BERT.
Following \citet{beltagy2020longformer}, we use a fully-connected layer on top of the first [CLS] token with global attention.
Longformer outperformed a RoBERTa baseline significantly on a small binary classification dataset \cite{beltagy2020longformer}. 
However, it has not been evaluated against any other models for text classification or on larger datasets that contain long documents.

\paragraph{ToBERT (chunk representations)} Transformer over BERT \cite[ToBERT,][]{pappagari2019hierarchical} takes a hierarchical approach that can process documents of any lengths in their entirety. 
The model divides long documents into smaller chunks of 200 tokens and uses a Transformer layer over BERT-based chunk representations.
It is reported to outperform previous state-of-the-art models on datasets of spoken conversations. However, it has not been compared to other Transformer models. 
We re-implement this model based on the specifications reported in \citet{pappagari2019hierarchical} as the code is not publicly available.

\paragraph{CogLTX (key sentence selection)} Cognize Long TeXts \cite[CogLTX,][]{ding_cogltx_2020} jointly trains two BERT (or RoBERTa) models to select key sentences from long documents for various tasks including text classification. 
The underlying idea that a few key sentences are sufficient for a given task has been explored for question answering \cite[e.g.][]{Min_2018}, but not much for text classification.
It is reported to outperform ToBERT and some other neural models (e.g. CNN), but it is not evaluated against other Transformer models. 

We use their multi-class classification code for any classification task with appropriate loss functions.\footnote{\url{https://github.com/Sleepychord/CogLTX}} 
Following \citet{beltagy2020longformer}, we use sigmoid and binary cross entropy loss on the logit output of the models for binary classification. The same setting is used for multi-label classification with softmax normalization and cross entropy loss. 

\subsection{Novel Baselines}
In addition to the representative models above, we include two novel methods that serve as simple but strong baseline models.

\paragraph{BERT+TextRank} While the BERT truncation baseline is often effective, key information required to classify documents is not always found within the first 512 tokens. To account for this, we augment the first 512 tokens, with a second set of 512 tokens obtained via TextRank, an efficient unsupervised sentence ranking algorithm \cite{mihalcea-tarau-2004-textrank}.
TextRank provides an efficient alternative to more complex models designed to select key sentences such as CogLTX.
Specifically, we concatenate the BERT representation of the first 512 tokens with that of the top ranked sentences from TextRank (up to another 512 tokens). As before, we use a fully-connected layer on top of the concatenated representation for classification.
We use PyTextRank \cite{PyTextRank} as part of the spaCy pipeline \cite{spacy} for the implementation with the default settings. 

\paragraph{BERT+Random} As an alternative approach to the BERT+TextRank model, we select random sentences up to 512 tokens to augment the first 512 tokens. 
Like BERT+TextRank, this can be a simple baseline approach in case key information is missing in truncated documents.\footnote{For simplicity, sentences included in the first 512 tokens are not excluded in the random selection process. Different settings are possible, but our preliminary results did not show much difference.}

\subsection{Hyperparameters}

We use reported hyperparameters for the existing models whenever available. However, given that we include different datasets that the original papers did not use, we additionally explore different hyperparameters for the models. Detailed information is available in Appendix~\ref{sec:appendix-hyper}.

\subsection{Data}

We select three classification datasets containing long documents to cover various kinds of classification tasks: Hyperpartisan \cite{kiesel-etal-2019-semeval} (binary classification), 20NewsGroups \cite{Lang95-20news} (multi-class classification) and EURLEX-57K  \cite{chalkidis-etal-2019-large} (multi-label classification).
We also re-purpose the CMU Book Summary Dataset \cite{bamman2013new} as an additional multi-label classification dataset. 

\begin{table}[t]
    \centering
    \begin{tabular}{lrrr}
        \hline
        Dataset & \# BERT Tokens & \% Long\\
        \hline
        \hline
        Hyperpartisan & 744.2 $\pm$ 677.9 & 53.5\\
        20NewsGroups & 368.8 $\pm$ 783.8 & 14.7\\
        EURLEX-57K & 
        707.99 $\pm$ 538.7 & 51.3\\
        Book Summary & 574.3 $\pm$ 659.6 & 38.8\\
        \hspace{10pt}-- Paired & 1,148.6 $\pm$ 933.9 & 75.5\\
        \hline
    \end{tabular}
    \caption{Statistics on the datasets. \# BERT Tokens refers to the average token count obtained via the tokenizer of the BERT base (uncased) model. \% Long refers to the percentage of documents with over 512 BERT tokens.} 
    \label{tab:datasets}
\end{table}

We also modify the EURLEX and Book Summary datasets to represent different data settings and further test all models under these challenging variations.
A document in the EURLEX dataset contains a legal text divided into several sections, and the first two sections (header, recitals) carry the most relevant information for classification \cite{chalkidis-etal-2019-large}. We invert the order of the sections so that this key information is located toward the end of each document (Inverted EURLEX). This creates a dataset particularly challenging for models that focus only on the first 512 tokens.
We also combine pairs of book summaries from the CMU Book Summary dataset to create a new dataset (Paired Book Summary) that contains longer documents with two distinctive information blocks. 
Again, this challenges models not to solely rely on the signals from the first 512 tokens.
In addition, it further challenges models to detect two separate sets of signals for correct classification results.
In all, these modified datasets represent different ways information may be presented in long texts and test how robust the existing models are to these. Table \ref{tab:datasets} summarizes characteristics of all our datasets, with more details in Appendix~\ref{sec:appendix-datasets}. 

\begin{table*}[ht!]
    \centering
    \begin{tabularx}{\textwidth}{l *{6}{Y} }
        \hline
        \multirow{2}{*}{Model}  & Hyper- & 20News & \multirow{2}{*}{EURLEX} & Inverted  & Book &Paired \\
        & partisan & Groups && EURLEX & Summary & Summary\\
        \hline
        \hline
        BERT & 92.00  & 84.79  & \underline{73.09}  & 70.53  & 58.18 & 52.24 \\
        BERT+TextRank & \cellcolor{Gray} 91.15 & \underline{84.99} & \cellcolor{Gray}72.87  &\underline{71.30}  &  \underline{58.94}  & 55.99\\
        BERT+Random & \cellcolor{Gray}89.23  & \cellcolor{Gray}84.65 & \textbf{73.22} & \textbf{71.47}& \textbf{59.36} & 56.58 \\
        Longformer & \textbf{95.69} & \cellcolor{Gray}83.39 & \cellcolor{Gray}54.53  &  \cellcolor{Gray}56.47 & \cellcolor{Gray}56.53  & \textbf{57.76} \\
        ToBERT & \cellcolor{Gray}89.54 & \textbf{85.52} &\cellcolor{Gray}67.57  & \cellcolor{Gray}67.31& \cellcolor{Gray}58.16 & \underline{57.08} \\
        CogLTX & \underline{94.77} & \cellcolor{Gray}84.63  & \cellcolor{Gray}70.13  & 70.80& 58.27  & 55.91 \\
        \hline
    \end{tabularx}
    \caption{Performance metrics on the test set for all datasets. The average accuracy (\%) over five runs is reported for Hyperpartisan and 20NewsGroups while the average micro-$F_{1}$ (\%) is used for the other datasets. The highest value per column is in bold and the second highest value is underlined. Results below the BERT baseline are shaded. 
    }
    \label{tab:results-on-all-docs}
\end{table*}

\subsection{Metrics}

For the binary (Hyperpartisan) and multi-class (20NewsGroups) classification tasks, we report accuracy (\%) on the test set. 
For the rest, multi-label classification datasets, we use micro-$F_{1}$ (\%), which is based on summing up the individual true positives, false positives, and false negatives for each class.\footnote{The choice of these metrics are based on previous literature. An exploration of other metrics (e.g. macro-$F_{1}$) may provide further insights. However, we did not see significant differences in preliminary results, and we believe the general trend of results would not differ.}

\section{Results}

Table \ref{tab:results-on-all-docs} summarizes the average performance of the models over five runs with different random seeds.
Overall, the key takeaway is that more sophisticated models (Longformer, ToBERT, CogLTX) do not outperform the baseline models across the board.
In fact, these models are significantly more accurate than the baselines only on two datasets.
As reported in \citet{beltagy2020longformer}, Longformer recorded the strongest performance on Hyperpartisan, with CogLTX also performing well.
Longformer and ToBERT performed the best for Paired Book Summary. 
Paired Book Summary seems to be most challenging for all models across the board and is the only dataset where the BERT baseline did the worst. However, it is worth noting that simple augmentations of the BERT baseline as in BERT+TextRank and BERT+Random were not far behind the best performing model even for this challenging dataset.
ToBERT's reported performance was the highest for 20NewsGroups, but we were unable to reproduce the results due to its memory constraints.
For the other datasets, these more sophisticated models were outperformed by the baselines. 
In particular, the simplest BERT baseline that truncates documents up to the first 512 tokens shows competitive performance overall, outperforming the majority of models for Hyperpartisan, 20NewsGroups and EURLEX. It is only the Paired Book Summary dataset where the BERT baseline performed particularly worse than other models.
In general, we observe little-to-no performance gains from more sophisticated models across the datasets as compared to simpler models.
A similar trend was observed even when the models were evaluated only on long documents in the test set (Appendix~\ref{sec:long-doc-results}). 
These finding suggests that the existing models do not necessarily work better for long documents across the board when diverse datasets are considered.

\begin{table}[t]
    \centering
    \begin{tabular}{l|rrc}
        \hline
        \multirow{2}{*}{Model} & Train & Inference & GPU\\
         &  Time &  Time &  Memory \\
        \hline
        \hline
        BERT & 1.00 & 1.00 & $<$16 \\ 
        +TextRank & 1.96 & 1.96 & 16\\ 
        +Random & 1.98 & 2.00 & 16\\
        Longformer & 12.05 & 11.92 & 32\\
        ToBERT & 1.19 & 1.70 & 32\\
        CogLTX & 104.52 & 12.53& $<$16\\
        \hline
    \end{tabular}
    \caption{Runtime and memory requirements of each model, relative to BERT, based on experiments on the Hyperpartisan dataset. 
    Training and inference time were measured and compared in seconds per epoch.
    GPU memory requirement is in GB. Longformer and ToBERT were trained on a GPU with a larger memory and compared to a comparable run on the machine.}
    \label{tab:time-space-results}
\end{table}

The relatively inconsistent performance of these existing models is even more underwhelming considering the difference in runtime and memory requirements as summarized in Table \ref{tab:time-space-results}. 
Compared to BERT on the first 512 tokens, 
Longformer takes about 12x more time for training and inference while CogLTX takes even longer.
ToBERT is faster than those two, but it requires much more GPU memory to process long documents in their entirety.
Taken together with the inconsistency in accuracy/F1 scores, this suggests that sophisticated models are not necessarily a good fit for real word use cases where efficiency is critical.

\section{Discussion and Recommendations}

Our results show that complex models for long document classification do not consistently outperform simple baselines.
The fact that the existing models were often outperformed by the simplest BERT baseline suggests that the datasets tend to have key information accessible in the first 512 tokens.
This is somewhat expected as the first two sections of EURLEX are reported to carry the most information \cite{chalkidis-etal-2019-large} and 20NewsGroups contains mostly short documents. 
Including these datasets to evaluate models for long document classification is still reasonable given that a good model should work well across different settings.
However, these datasets alone do not represent various ways information is presented in long texts.

Instead, future studies should evaluate their models across various datasets to create robust models.
While it is often difficult to obtain datasets suited for long document classification, our modifications of existing datasets may provide ways to repurpose existing datasets for future studies.
We invert the order of the sections of EURLEX to create the Inverted EURLEX dataset, where key information is likely to appear toward the end of each document. 
Our results in Table \ref{tab:results-on-all-docs} show that selective models (BERT+TextRank, BERT+Random, CogLTX) performed better than those that read longer consecutive sequences (Longformer, ToBERT) on this dataset. 
This suggests that this inverted dataset may contain parts of texts that should be ignored for better performance, thus providing a novel test bed for future studies.
The Paired Book Summary dataset presents another challenging data setting with two distinctive information blocks.
While Longformer and ToBERT performed significantly better for this dataset than others, the overall model performance was quite underwhelming, leaving room for improvement for future models.

Many of these findings were revealed only due to the choice of  relevant baselines, and future work will benefit from including these as well. 
A BERT/RoBERTa baseline is essential to motivate the problem of long document classification using Transformers and reveal how much information is retrievable in the first 512 tokens. 
BERT+TextRank and BERT+Random are stronger baselines that often outperform more complex models that select key sentences.
In fact, they outperformed CogLTX on five of the six datasets.

\section{Conclusion}

Several approaches have been proposed to use Transformers to classify long documents, yet their relative efficacy remains unknown. 
In this paper, we compare existing models and baselines on various datasets and in terms of their time and space requirements.
Our results show that existing models, while requiring more time and/or space, do not perform consistently well across datasets, and are often outperformed by baseline models. 
Future studies should consider the baselines and datasets to establish robust performance.

\section*{Acknowledgments}
We would like to thank the reviewers and area chairs for their thoughtful comments and suggestions. We also thank the members of AWS AI Labs for many useful discussions and feedback that shaped this work.

\bibliography{long-doc}
\bibliographystyle{acl_natbib}

\appendix

\begin{table*}[ht!]
    \centering
    \begin{tabular}{l|rrrrrrr}
        \hline
        Dataset & Type & \# Train & \# Dev & \# Test & \# Labels & \# BERT Tokens & \% Long\\
        \hline
        \hline
        Hyperpartisan & binary & 516 & 64 & 65 & 2 & 744.18 $\pm$ 677.87 & 53.49\\
        20NewsGroups & multi-class & 10,182 & 1,132 & 7,532 & 20 & 368.83 $\pm$ 783.84 & 14.71\\
        EURLEX-57K & \multirow{2}{*}{multi-label} & \multirow{2}{*}{45,000} & \multirow{2}{*}{6,000} & \multirow{2}{*}{6,000} & \multirow{2}{*}{4,271} & \multirow{2}{*}{707.99 $\pm$ 538.69} & \multirow{2}{*}{51.30}\\
        \hspace{10pt}-- Inverted & &  &   &  &  &  &  \\
        Book Summary & multi-label & 10,230 & 1,279 & 1,279 & 227 & 574.31 $\pm$ 659.56 & 38.76\\
        \hspace{10pt}-- Paired & multi-label & 5,115 & 639 & 639 & 227 & 1,148.62 $\pm$ 933.97 & 75.54\\
        \hline
    \end{tabular}
    \caption{Statistics on the datasets. 
    \# BERT Tokens refers to the average token count obtained via the tokenizer of the BERT base model (uncased).
    \% Long refers to the percentage of documents with more than 512 BERT tokens.} 
    \label{tab:datasets-2}
\end{table*}

\begin{table*}[ht]
    \centering
    \begin{tabularx}{\textwidth}{l| *{6}{Y} }
        \hline
        \multirow{2}{*}{Model}  & Hyper- & 20News & \multirow{2}{*}{EURLEX} & Inverted  & Book &Paired \\
        & partisan & Groups && EURLEX & Summary & Summary\\
        \hline
        \hline
        BERT & 88.00& \underline{86.09}& \underline{66.76}& 62.88 & 60.56  & 52.23 \\
        BERT+TextRank & \cellcolor{Gray}85.63 & \cellcolor{Gray}85.55 & \cellcolor{Gray}66.56 & \underline{64.22}& \underline{61.76}   & 56.24\\
        BERT+Random & \cellcolor{Gray}83.50& \textbf{86.18} & \textbf{67.03}  & \textbf{64.31}& \textbf{62.34} & 56.77\\
        Longformer & \textbf{93.17}  & \cellcolor{Gray}85.50& \cellcolor{Gray}44.66 & \cellcolor{Gray}47.00& \cellcolor{Gray}59.66
        & \textbf{58.85}\\
        ToBERT & \cellcolor{Gray}86.50 &\,\,\,\, -- \,\,\,\, & \cellcolor{Gray}61.85& \cellcolor{Gray}59.50 & 61.38
         & \underline{58.17}\\
        CogLTX & \underline{91.91}& \cellcolor{Gray}86.07& \cellcolor{Gray}61.95  & 63.00 & 60.71& 55.74\\
        \hline
    \end{tabularx}
    \caption{Performance metrics evaluated on long documents in the test set for all datasets. The average accuracy (\%) over five runs is reported for Hyperpartisan and 20NewsGroups while the average micro-$F_{1}$ (\%) is used for the other datasets. The highest value per column is in bold and the second highest value is underlined. Results below the BERT baseline are shaded. Running ToBERT on 20NewsGroups seems to require further preprocessing, which we were unable to replicate with the reported information.}
    \label{tab:results-on-long-docs}
\end{table*}

\section{Hyperparameters}
\label{sec:appendix-hyper}

Across all datasets, we used Adam optimizer with a learning rate of \{5e-5, 3e-5, 0.005\} for one run of each model and picked the best performing learning rate for the model. The learning rate of 0.005 was used for Longformer only because it did not perform well with a learning rate of 5e-5 or 3e-5 for most of the datasets.
We set dropout rate at 0.1 as suggested by \citet{devlin-etal-2019-bert}.
The number of epochs needed for finetuning the models for different datasets is likely to vary, so we trained all models for 20 epochs and selected the best performing model based on the performance metric on the validation set.
We report the average results on the test set over five different seeds.

All experiments on baseline models and CogLTX were conducted on a single Tesla V100 GPU with 16GB memory.
For Longformer and ToBERT, we used a NVIDIA A100 SXM4 with 40GB memory. 
More details on the selected hyperparameters are available with our code at \url{https://github.com/amazon-research/efficient-longdoc-classification}.

\section{Datasets}
\label{sec:appendix-datasets}

\paragraph{Hyperpartisan} is a binary classification dataset, where each article is labeled as True (hyperpartisan) or False (not hyperpartisan) \cite{kiesel-etal-2019-semeval}. More than half of the documents exceed 512 tokens. It is quite different from other datasets in that it is a very small dataset: the training set contains 516 documents while the development and test sets contain 64 and 65 documents, respectively.

\paragraph{20NewsGroups} is a widely-used multi-class classification dataset \cite{Lang95-20news}. The documents are categorized into well-balanced, 20 classes. Only about 15\% of the documents exceed 512 tokens. 
While the original dataset comes in train and test sets only, we report results on the train/dev/test split as used in \citet{pappagari2019hierarchical}, where we take 10\% of the original train set as the development set.
Note that CogLTX reported their accuracy at 87.00\% on the test set and 87.40\% on the long documents in the test set, using the original train and test sets only. 
Our implementation of CogLTX in the same setting with five different runs resulted in a much lower performance at 85.15\% on the test set and 86.57\% on the long documents only.
In addition, we were unable to replicate ToBERT results on 20NewsGroups.
It is unclear how the dataset is further preporcessed for ToBERT, and our implementation of ToBERT caused a GPU out-of-memory error on 20NewsGroups. Thus, we show the reported results for ToBERT on this dataset. 

\paragraph{EURLEX-57K} is a multi-label classification dataset based on EU legal documents \cite{chalkidis-etal-2019-large}. In total, there are 4,271 labels available, and some of them do not appear in the training set often or at all, making it a very challenging dataset. About half of the datasets are long documents. Each document contains four major zones: header, recitals, main body, and attachments. \citet{chalkidis-etal-2019-large} observe that processing the first two sections only (header and recitals) results in almost the same performance as the full documents and that BERT on the first 512 tokens outperforms all the other models they considered. After examining the dataset, we exclude the attachments section as it does not seem to provide much textual information.

\paragraph{CMU Book Summary} contains book summaries extracted from Wikipedia with corresponding metadata from Freebase such as the book author and genre \cite{bamman2013new}. We use the summaries and their corresponding genres for a multi-label classification task. We keep 12,788 out of 16,559 documents after removing data points missing any genre information and/or adequate summary information (e.g. less than 10 words). In total, there are 227 genre labels such as `Fiction' and `Children's literature'.

\section{Results on long documents only}
\label{sec:long-doc-results}

Table \ref{tab:results-on-long-docs} shows the results as evaluated on long documents (with over 512 tokens) in the test set only. 
Overall, the results show a similar trend as observed in Table \ref{tab:results-on-all-docs}, which reports the results on the entire documents in the test set.
In general, the existing models were often outperformed by the BERT truncation baseline.
This suggest that these models designed for long document classification do not perform particularly well on the long documents in the datasets.
The only difference is that BERT+Random and ToBERT perform better than the BERT baseline when evaluated on long documents only for 20NewsGroups and Book Summary, respectively.
However, the performance gain does not seem significant, and the relative performance with respect to the other models remains largely unchanged. 
In general, the relative strength of a model for a given dataset stays the same whether or not the model is evaluated on the entire documents or long documents in the test set.

\end{document}